\def\year{2021}\relax
\documentclass[letterpaper]{article}
\usepackage{aaai21} 
\usepackage{bm}
\usepackage{color}
\usepackage{amssymb}
\usepackage{amsmath}
\usepackage{multirow}
\usepackage{times}  
\usepackage{helvet} 
\usepackage{courier}
\usepackage[hyphens]{url} 
\usepackage{graphicx} 
\usepackage{natbib}
\usepackage{caption}
\usepackage{pifont}

\newcommand{\cmark}{\text{\ding{51}}}

\title{Task-Independent Knowledge Makes for Transferable Representations for Generalized Zero-Shot Learning}

\author{
	Chaoqun Wang\textsuperscript{\rm 1 \rm 2}, Xuejin Chen\footnote{Corresponding Author}\textsuperscript{\rm 1 \rm 2}, Shaobo Min\textsuperscript{\rm 2}, Xiaoyan Sun\textsuperscript{\rm 2}, Houqiang Li\textsuperscript{\rm 1 \rm 2}%  
}
\affiliations{
	
	\textsuperscript{\rm 1}School of Data Science
	
	\textsuperscript{\rm 2}The National Engineering Laboratory for Brain-inspired Intelligence Technology and Application
	
	University of Science and Technology of China, Hefei, Anhui, China\\
	
	cq14@mail.ustc.edu.cn, xjchen99@ustc.edu.cn, mbobo@mail.ustc.edu.cn, \{sunxiaoyan, lihq\}@ustc.edu.cn

}

\begin{document}

\maketitle

\begin{abstract}
Generalized Zero-Shot Learning (GZSL) targets recognizing new categories by learning transferable image representations.
Existing methods find that, by aligning image representations with corresponding semantic labels, the semantic-aligned representations can be transferred to unseen categories.
However, supervised by only seen category labels, the learned semantic knowledge is highly task-specific, which makes image representations biased towards seen categories.
In this paper, we propose a novel Dual-Contrastive Embedding Network (DCEN) that simultaneously learns task-specific and task-independent knowledge via semantic alignment and instance discrimination.
First, DCEN leverages task labels to cluster representations of the same semantic category by cross-modal contrastive learning and exploring semantic-visual complementarity.
Besides task-specific knowledge, DCEN then introduces task-independent knowledge by attracting representations of different views of the same image and repelling representations of different images.
Compared to high-level seen category supervision, this instance discrimination supervision encourages DCEN to capture low-level visual knowledge, which is less biased toward seen categories and alleviates the representation bias.
Consequently, the task-specific and task-independent knowledge jointly make for transferable representations of DCEN, which obtains averaged $4.1\%$ improvement on four public benchmarks.

\end{abstract}

\section{Introduction}
Deep learning-based methods are highly successful on various computer vision tasks, such as image classification \cite{He_2016_CVPR}, object detection \cite{girshick2014rich}, and semantic segmentation \cite{badrinarayanan2017segnet}. However, these models require a huge demand for manually labelled training data for numerous classes.  
To this end, Generalized Zero-Shot Learning (GZSL), which aims to recognize either seen or unseen categories thereby reduces manual annotation labors, recently has attracted great interests.

\begin{figure}[t!]
	\centering
	\includegraphics[width=0.95\columnwidth]{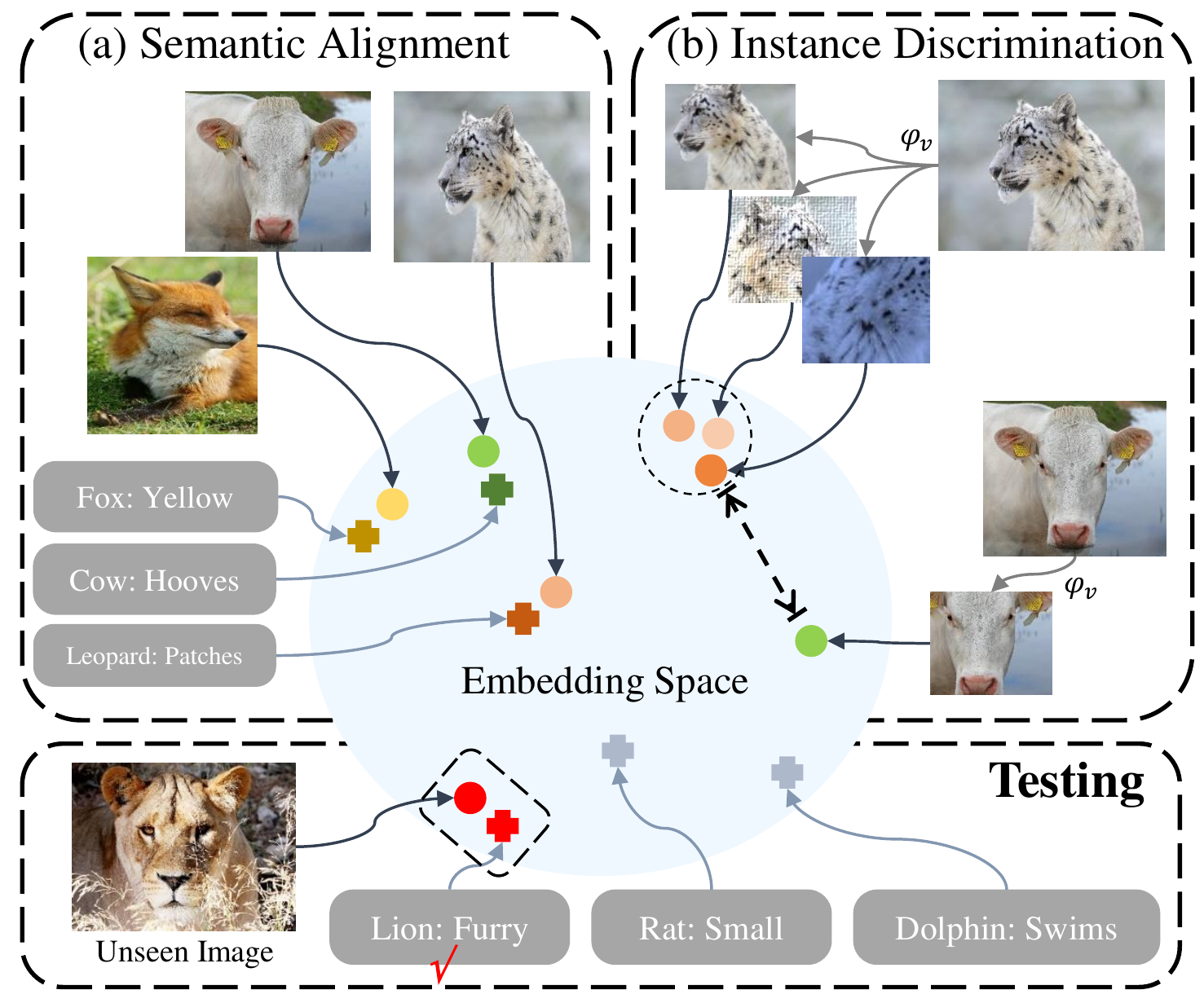} 
	\caption{Motivation of this paper. (a) Existing methods focus on using task labels to learn semantic-aligned representations, which can be transferred to unseen categories. (b) Besides, this paper further learns task-independent knowledge via instance discrimination supervision, which significantly improves the representation transferability.}
	
	\label{fig:moti}
	\vspace{-0.5cm} 
\end{figure}

Due to unavailable unseen category data during training, GZSL puts a high demand on transferability of image representations \cite{Jiang_2018_ECCV,tong2019hierarchical}.
Previous methods tackle this problem by aligning image representations with corresponding category attributes \cite{xian2018zero}, as shown in Fig.~\ref{fig:moti} (a).
Since seen and unseen domain categories share a common attribute space, the semantic-aligned representation can be transferred to unseen categories, and the recognition becomes a nearest neighbour searching problem, \emph{i.e.,} images and category attributes serve as queries and anchors, respectively.
Based on this paradigm, most GZSL methods focus on designing elaborate semantic-visual alignment mechanism, such as visual and attribute attentions \cite{zhu2019semantic,xie2019attentive,Huynh_2020_CVPR}, semantic auto-encoders \cite{felix2018multi,min2019domain}, and domain detectors \cite{atzmon2019adaptive}.
However, trained with only seen category labels, the learned image representations are highly task-specific and biased towards seen categories, which makes an unseen image tend to be recognized as seen categories.

To alleviate this issue, we introduce task-independent knowledge that is learned without seen category concepts and shared by both seen and unseen images.
We propose a novel Dual-Contrastive Embedding Network (DCEN) that simultaneously learns task-specific and task-independent knowledge to obtain transferable representations in GZSL.
Specifically, we design two modules in DCEN,~\emph{i.e.,} Semantic Contrastive Module (SCM) and Visual Contrastive Module (VCM), to enhance the semantic-visual alignment knowledge via task labels and introduce annotation-free visual knowledge via instance discrimination,  respectively. 

First, SCM is developed to leverage semantic labels to learn task-specific knowledge.
By constructing cross-modal triplets, SCM constrains the image representation to be not only aligned with the corresponding category but also far away from the most confusing category, which improves the inter-class representation discrimination.
Besides, we design masked attribute prediction to complete the missing category attributes from corresponding image representations.
This renders the image representation to preserve the semantic-visual complementarity, resulting in better semantic-visual alignment.
With cross-modal contrastive learning and masked attribute prediction, SCM can better bridge the semantic-visual gap than previous methods to learn semantic-aligned representation.

Second, VCM is designed to learn task-independent knowledge by distinguishing each image as an individual category, as shown in Fig.~\ref{fig:moti} (b).
This forces the image representation to capture as many detailed visual cues as possible for each individual image, so that all images can be separated apart.
Without task labels, the learned low-level visual knowledge is only related to intra-image invariance and inter-image difference, which is irrelevant to seen category concepts.
Thus, the learned knowledge is less biased towards the seen categories than the task-specific knowledge from seen category labels, and thereby significantly improves the representation transferability.

Consequently, the task-independent knowledge from instance discrimination and task-specific knowledge from semantic alignment jointly make for transferable representations of our DCEN, which achieves averaged $4.1\%$ improvement on four widely-used GZSL benchmarks.
Our overall contribution is threefold:
\begin{itemize}
\setlength{\itemsep}{0pt}
\setlength{\parsep}{0pt}
\setlength{\parskip}{0pt}

\item We propose a novel Dual-Contrastive Embedding Network (DCEN) that simultaneously learns task-specific and task-independent knowledge to produce transferable representations in GZSL.
To the best of our knowledge, this is the first work that introduces task-independent knowledge from instance discrimination to boost representation transferability in GZSL.

\item DCEN learns task-independent knowledge by exploring inter-image visual distinction and intra-image transformation invariance. Extensive experiments are conducted to search important low-level invariance concepts for GZSL.

\item DCEN learns task-specific knowledge by constructing cross-modal contrastive learning and exploring semantic-visual complementarity to better bridge the semantic-visual gap.

\end{itemize}

\section{Related Works}
\subsection{Generalized Zero-Shot Learning}
This paper belongs to Inductive GZSL \cite{Xian2016,kodirov2017semantic,min2020domain}, where the unseen domain data is unavailable during training. 
A general solution is to learn a joint embedding space, where the image representations and semantic labels,~\emph{e.g.,} category attributes \cite{farhadi2009describing,Morgado2017} or text descriptions \cite{lei2015predicting}, are aligned \cite{xian2018zero,zhu2019generalized}.

Based on this motivation, most existing methods focus on boosting the semantic-visual alignment by carefully designing the embedding space.
For example, previous methods \cite{Shigeto2015,zhu2019generalized,min2019domain} constrain the embedding space to be spanned by high-dimensional visual representations, instead of semantic labels, to improve the space discrimination.
This can prevent a few seen categories from being the anchors for most input images \cite{Annadani2018}, but it also weakens the semantic relationship in the embedding space.
To this end, recent methods \cite{kodirov2017semantic,tong2019hierarchical} use auto-encoders to enhance the semantic relationship in the embedding space.
For example, DSEN \cite{min2019domain} designs domain-specific cyclic encoders between semantic embeddings and category attributes to explore semantic domain differences, and SP-AEN \cite{Chen2018} uses cyclic constraints between visual representations and images to capture detailed visual clues.
Recently, attention mechanisms are leveraged for images and semantic labels to improve representation discrimination.
In VSE \cite{zhu2019generalized} and SGMA \cite{zhu2019semantic}, multi-head attentions are designed to infer important local regions to boost semantic-visual alignment. LFGAA \cite{Liu_2019_ICCV}, and DAZLE \cite{Huynh_2020_CVPR} use attribute attention to generate clean semantic embeddings.
Although being effective, the previous methods focus on better leveraging seen domain labels to improve semantic-visual alignment.
Thus, it inevitably makes the image representation highly task-specific and biased towards seen categories, which limits the transferability to unseen categories.

\begin{figure*}[t!]
	\centering
	\includegraphics[width=1.9\columnwidth]{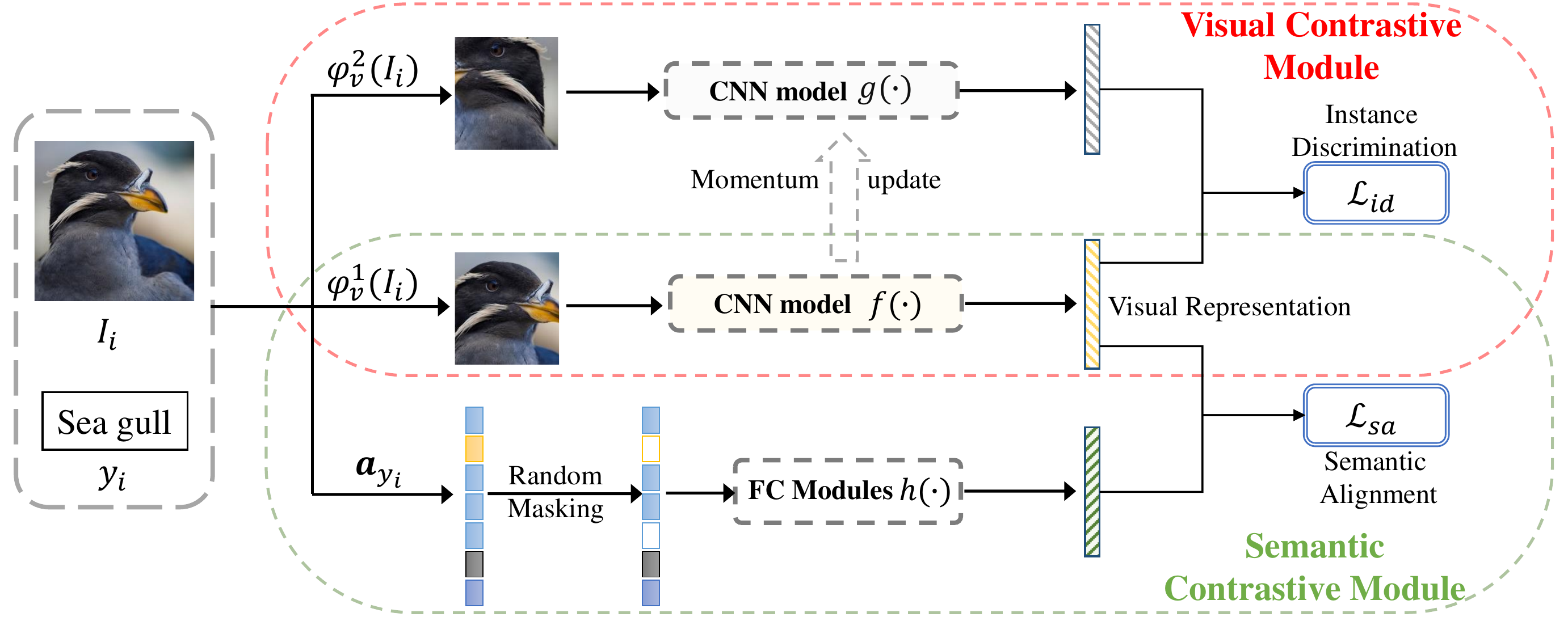} 
	\caption{The diagram of Dual-Contrastive Embedding Network. Visual Contrastive Module learns task-independent knowledge via instance discrimination, and Semantic  Contrastive Module learns task-specific knowledge via semantic alignment.}
	\label{fig:pipeline}
	
\end{figure*}

\subsection{Self-Supervised Learning}
Without human annotations, self-supervised learning methods \cite{tian2020contrastive} have proved that learning task-independent knowledge can produce representations with strong generalization on various downstream tasks, which even surpass fully-supervised methods.
Specifically, they usually construct some proxy tasks as supervision to learn image representations.
For example, RotNet \cite{gidaris2018unsupervised} constrains the model to predict the rotation angles of an input image. 
CFN \cite{noroozi2016unsupervised} regards an image as a jigsaw and predicts the shuffled patch order. Deep clustering methods \cite{yang2018new,caron2018deep,yang2020adversarial} separate different classes via clustering.
Recently, contrastive learning \cite{tian2020contrastive,He_2020_CVPR} has attracted increasing attention, which aims to minimize the mutual information between different images and regards each image as an individual category.
Some works extend contrastive learning to cross-modal tasks. AVID \cite{morgado2020audio} utilizes the correspondence between audio and visual representations. CMC \cite{tian2020contrastive} contrasts across multiple views, such as luminance, chrominance RGB, and depth.
Inspired by the generalized self-supervised representation, task-independent knowledge has been introduced in many other fields, such as few-shot learning \cite{gidaris2019boosting} and landmark detection \cite{cheng2020unsupervised}.

In this paper, we explore the task-independent knowledge from instance discrimination to boost representation transferability in GZSL. 
Notably, SDGN \cite{wu2020self} also uses self-supervised learning in GZSL.
However, it is much different from our DCEN, because: (a) SDGN follows Transductive ZSL setting that uses unseen domain images for training, while our DCEN only uses data from seen domain during the training stage; (b) SDGN applies contrastive loss to triplet data of two domains to alleviate domain confusion, 
while we utilize instance discrimination knowledge of the seen domain; (c) SDGN is a generative method that utilizes powerful GANs for unseen domain feature generation.

\section{Dual-Contrastive Embedding Network}

\subsection{Problem Formulation}
The target of GZSL is to recognize images from either seen or unseen categories using the model trained only with seen domain data.
Here, we define seen domain data as 
$\mathcal{S}=\{I_i,y_i,\boldsymbol{a}_{y_i}|I_i\in\mathcal{I}_s,y_i\in \mathcal{Y}_s,\boldsymbol{a}_{y_i}\in \mathcal{A}_s \}$, where $I_i$ is an image of a seen category, $y_i$ is the corresponding category, and $\boldsymbol{a}_{y_i}$ is the semantic description of $y_i$, such as category attributes.
Unseen domain data are similarly defined as 
$\mathcal{U}=\{I_i,y_i,\boldsymbol{a}_{y_i}|I_i\in\mathcal{I}_u,y_i\in \mathcal{Y}_u,\boldsymbol{a}_{y_i}\in \mathcal{A}_u \}$, where $\mathcal{Y}_s \cap\mathcal{Y}_u=\emptyset$.

One main stream of GZSL methods is to learn a joint embedding space, where the image representations and category descriptions can be aligned by minimizing: 
\begin{equation}
\label{eq:L_zsl}
\mathcal{L}_{zsl}=\sum_{I_i\in\mathcal{I}_s}d\big(f(I_i),\boldsymbol{a}_{y_i}\big),
\end{equation}
where 
$f(\cdot)$ is the feature extractor to produce image representations. 
$d(\cdot,\cdot)$ is a distance function, such as negative cosine similarity. 
Since $\mathcal{A}_s$ and $\mathcal{A}_u$ usually share a common semantic space, the semantic-visual relationship captured by $f(\cdot)$ can be transferred to the unseen domain by:

\begin{equation}
\label{eq:infer}
\hat{y}=\arg\min_{y\in\mathcal{Y}_s\cup \mathcal{Y}_u} d\big(f(I),\boldsymbol{a}_y\big),
\end{equation}
where $I\in\mathcal{I}_s\cup \mathcal{I}_u$.

However, due to unavailable $\mathcal{U}$ during training, it puts a high demand on transferability of learned $f(\cdot)$ in Eq.~\eqref{eq:L_zsl}.  

\subsection{Semantic Contrastive Module}
In order to better align semantic and visual embeddings, we introduce a Semantic Contrastive Module (SCM) to better bridge the semantic-visual gap via cross-modal contrastive learning and exploring semantic-visual complementarity.
Specifically, in the SCM, we construct semantic-visual triplets to prevent image representations from being biased towards confusing category and design masked attribute prediction to complete the missing attributes from image representations.

As shown in Fig.~\ref{fig:pipeline}, given an image $I_i$ and its category attribute $\boldsymbol{a}_{y_i}$, SCM learns the semantic-aligned representation $f(I_{i})$ by minimizing:
\begin{equation}
\label{eq:L_sa}
\mathcal{L}_{sa}=\sum_{I_i\in\mathcal{I}_s}d\big(f(I_{i}),h(\widehat{\boldsymbol{a}}_{y_i})\big)-\min\limits_{y_k\neq y_i}\{d\big(f(I_{i}),h(\widehat{\boldsymbol{a}}_{y_k})\big)\}.
\end{equation}
Different from the general objective function Eq.~\eqref{eq:L_zsl}, 
SCM first randomly masks some elements of $\boldsymbol{a}_{y_i}$,
which generates $\widehat{\boldsymbol{a}}_{y_i}$ to introduce some semantic variance and improve model robustness to input noises.
As attributes contain abstract semantic information, such as wing colors, it is hard for visual feature extractor $f(\cdot)$ to extract such highly abstract concepts from the images. Thus, instead of directly aligning $f(I_{i})$ with $\widehat{\boldsymbol{a}}_{y_i}$, SCM uses $K$ fully-connected (FC) modules, consisting of FC, Batch normalization (BN), and ReLU, to produce high-level semantic embeddings $h(\widehat{\boldsymbol{a}}_{y_i})$, which can better bridge the semantic-visual gap.
Besides, SCM constructs cross-modal triplet $\{f(I_{i}),h(\widehat{\boldsymbol{a}}_{y_i}),h(\widehat{\boldsymbol{a}}_{y_k})\}$, where $f(I_{i})$ is the anchor image representation, $h(\widehat{\boldsymbol{a}}_{y_i})$ is the positive category semantic embedding, and $h(\widehat{\boldsymbol{a}}_{y_k})$ is the most confusing negative category semantic embedding. Triplet loss has been demonstrated useful in previous works of face recognition \cite{schroff2015facenet}.
Thus, $\mathcal{L}_{sa}$ can not only align $f(I_{i})$ with the corresponding $h(\widehat{\boldsymbol{a}}_{y_i})$, but also punish the most confusing category.
Compared to $\mathcal{L}_{zsl}$, $\mathcal{L}_{sa}$ can produce more discriminative semantic-aligned $f(I_{i})$ via cross-modal contrastive learning.

\begin{figure}[t]
	\centering
	\includegraphics[width=1\columnwidth]{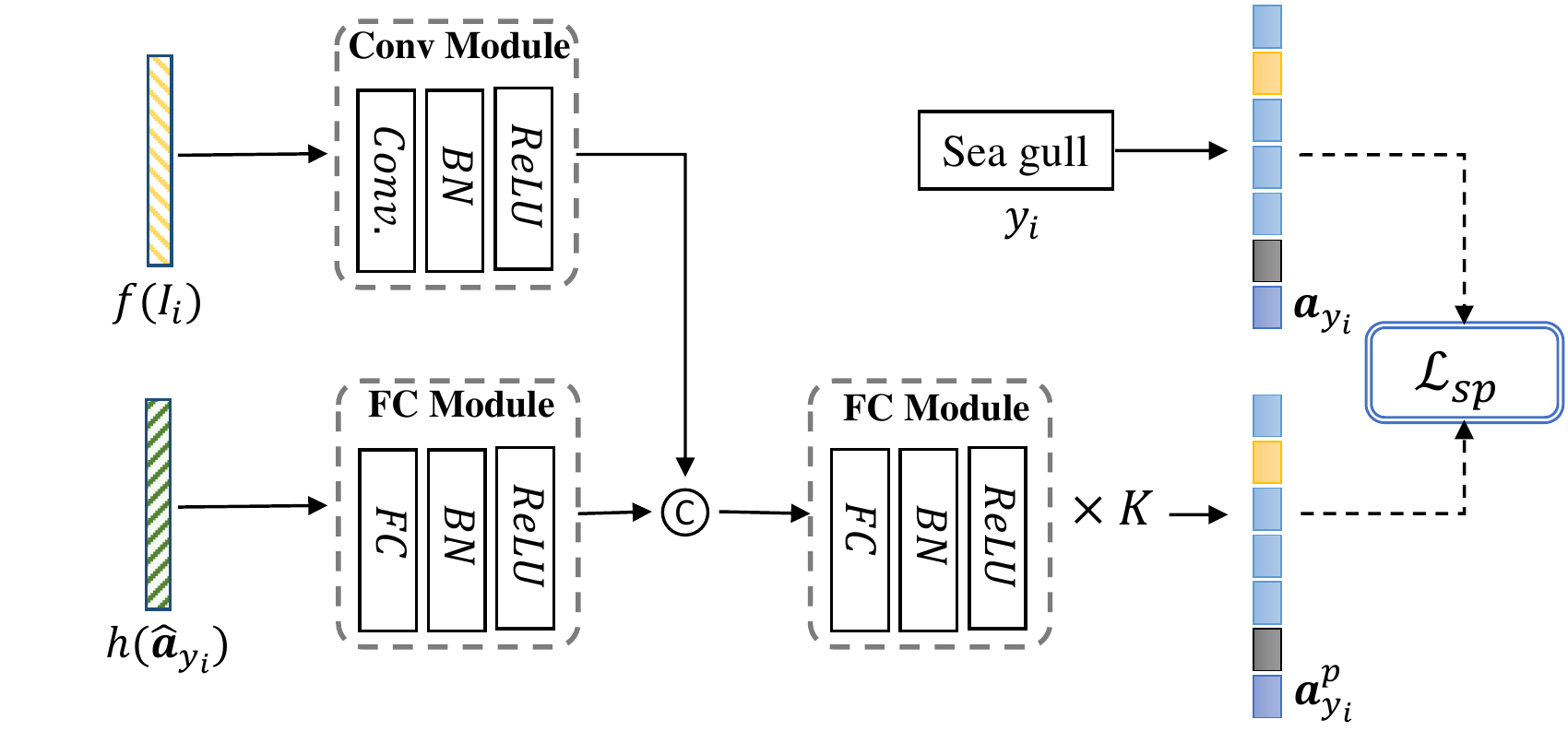} 
	\caption{Architecture for masked attribute prediction.}
		\vspace{-0.5cm} 
	\label{fig:vcsc}
\end{figure}

To better preserve semantic relationships among categories and encode the semantic knowledge into $f(I_i)$, masked attribute prediction $\widehat{h}(\cdot)$ is designed by predicting the missed elements in $\widehat{\boldsymbol{a}}_{y_i}$ from the image representation $f(I_i)$. The architecture is shown in Fig.~\ref{fig:vcsc}. 
we first fuse the image representation $f(I_i)$ and semantic embedding $h(\widehat{\boldsymbol{a}}_{y_i})$ via FC blocks and concatenation.
Then, a decoder module, consisting of $K$ FC blocks, is used to predict the intact attribute $\boldsymbol{a}^p_{y_i}$. The depths of $h(\cdot)$ and $\widehat{h}(\cdot)$ are kept consistent for better attribute reconstruction.
Notably, since the attribute masking is random and $h(\widehat{\boldsymbol{a}}_{y_i})$ has no masking knowledge, $f(I_i)$ should capture accurate and entire semantic information from $I_i$ to complete $\widehat{\boldsymbol{a}}_{y_i}$, thereby enhancing the semantic-visual alignment.
The loss function for masked attribute prediction is: 
\begin{equation}
\label{eq:L_sp}
\mathcal{L}_{sp}=\sum_{I_i\in\mathcal{I}_s}\left\|\widehat{h}\big(f(I_i),h(\widehat{\boldsymbol{a}}_{y_i})\big)-\boldsymbol{a}_{y_i}\right\|_2.
\end{equation}

With cross-modal contrastive learning and masked attribute prediction, SCM learns better semantic alignment knowledge than naive $\mathcal{L}_{zsl}$, which makes image representation $f(\cdot)$ better transferable to unseen categories.

\subsection{Visual Contrastive Module}
Since only $\mathcal{S}$ is available for training, the learned semantic knowledge in $f(\cdot)$ from SCM is inevitably task-specific to the seen categories.
Thus, a visual contrastive module (VCM) is proposed to learn task-independent knowledge without human annotations, which further improves the transferability of $f(\cdot)$.
To this end, VCM regards each image $I$ as an individual category, to push different image representations far away from each other.
This enforces $f(\cdot)$ to capture more detailed visual clues to represent each individual image so that they can be separated apart well.
Besides, VCM learns low-level visual invariance concepts, such as rotation and cropping invariance, that lead to strong generalization.

Given the $i$-th image $I_i$ in $\mathcal{I}_s$,
VCM first transforms  $I_i$ into two different counterparts $I_{i}^{v1}=\varphi_v^1(I_i)$ and $I_{i}^{v2}=\varphi_v^2(I_i)$.
$\varphi_v(\cdot)$ is a composition of random low-level transformations,~\emph{e.g.}, cropping, rotation, blurring, or color jittering. 
Then, $I_{i}^{v1}$ and $I_{i}^{v2}$ are fed into two CNN models $f(\cdot)$ and $g(\cdot)$ to extract visual representations $f(I_{i}^{v1})$ and $g(I_{i}^{v2})$, respectively. 
Finally, a visual contrastive loss is used to attract representations of two counterparts generated from the same image and repel representations from different images by:
\begin{equation}
\label{eq:L_id}
\mathcal{L}_{id}=-\sum_{I_i\in\mathcal{I}_s}\log\frac{\exp\{d(f(I_{i}^{v1}),g(I_{i}^{v2}))/\tau\}}{\sum_j\exp\{d(f(I_{i}^{v1}),g(I_{j}^{v2})) /\tau\}},
\end{equation}
where $d(\cdot,\cdot)$ is the cosine similarity between two representations, and $\tau$ is a temperature parameter. 
The positive data pair is defined as \{$I_{i}^{v1}$, $I_{i}^{v2}$\} that are two views from the same image, while the negative data pair is defined as \{$I_{i}^{v1}$, $I_{j}^{v2}$\} for different images when $i\neq j$. 
Data from previous batches serve as negative data. 
The numerator term of $\mathcal{L}_{id}$ constrains $f(\cdot)$ to be invariant to pre-defined image transformations $\varphi_v(\cdot)$.
By designing different image transformations $\varphi_v(\cdot)$, VCM can induce different low-level invariance concepts that are useful in GZSL tasks.
The denominator term of $\mathcal{L}_{id}$ constrains $f(\cdot)$ to be discriminative among different images.

In $\mathcal{L}_{id}$, the weights of $g(\cdot)$ are momentum-updated \cite{laine2016temporal} from $f(\cdot)$ by:
\begin{equation}
\label{eq:g_upate}
W_g = m*W_g+(1-m)*W_f,
\end{equation}
where $W_g$ and $W_f$ are weights of $g(\cdot)$ and $f(\cdot)$.
$m$ is a momentum parameter.
The momentum-updated $g(\cdot)$ has two main advantages: a) compared to a new trainable CNN, $g(\cdot)$ has less memory cost, and b) 
compared to sharing weights of $f(\cdot)$, $g(\cdot)$ is the temporal ensemble of different checkpoints from $f(\cdot)$ and contains temporal momentum information. 

Consequently, through instance discrimination $\mathcal{L}_{id}$, $f(\cdot)$ can gather augmented views of the same image and push different images apart. 
Thus, $f(\cdot)$ can not only capture the unique information contained in each individual image but also learn useful invariance concepts.
More importantly, this low-level visual knowledge is irrelevant to any category concept of $\mathcal{Y}_s$, thereby being well-transferable to the unseen images.

\subsection{Overall Objective}
Finally, the overall objective function of DCEN is :
\begin{equation}
\label{eq:L_obj}
\mathcal{L}_{all}=\lambda_1\mathcal{L}_{id}+\mathcal{L}_{sa}+\lambda_2\mathcal{L}_{sp},
\end{equation}
where $\lambda_1$ and $\lambda_2$ are hyper-parameters to balance $\mathcal{L}_{id}$ and $\mathcal{L}_{sp}$. 
The task-specific knowledge of $\mathcal{L}_{sa}+\lambda_2\mathcal{L}_{sp}$ and task-independent knowledge of $\lambda_1\mathcal{L}_{id}$ jointly make for transferable visual representation $f(I)$.

The testing phase is viewed as a nearest neighbour searching process by: 
\begin{equation}
\label{eq:infer1}
\hat{y}=\arg\min_{y\in\mathcal{Y}_s\cup \mathcal{Y}_u} d\big(f(I),h(\boldsymbol{a}_y)\big),
\end{equation}
where $I\in\mathcal{I}_s\cup \mathcal{I}_u$ is the query image,
and $\boldsymbol{a}_y$ is the anchor category attribute from both the seen and  unseen domains.

\section{Experiments}
In this section, we first evaluate each component of DCEN and then compare DCEN with state-of-the-art GZSL methods. Please refer to the supplementary for implementation details of DCEN.
\subsection{Experimental Settings}

\subsubsection{Datasets.}
We adopt four widely-used GZSL benchmarks, which are Caltech-USCD Birds-200-2011 (CUB) \cite{Welinder2010}, SUN \cite{patterson2012sun}, Animals with Attributes2 (AWA2) \cite{xian2018zero}, and Attribute Pascal and Yahoo (aPY) \cite{farhadi2009describing} for the following experiments. Detailed statistics of datasets are listed in Table \ref{tab:dataset}.
Category attributes provided by datasets are used as semantic labels.

\begin{table}
	\centering
	
	\resizebox{\columnwidth}{!}{
	\begin{tabular}{c|c|c|c|c|c}
		\hline
		Dataset & Seen/Unseen & Attributes & Train &Val & Test \\ \hline
		CUB & 150/50 & 312 & 7,057 & 1,764 & 2,967 \\ \hline
		AWA2 & 40/10 & 85 & 23,527  & 5,882 & 7,913 \\ \hline
		aPY & 20/12 & 64 & 5,932 & 1,483 & 7,924 \\ \hline
		SUN & 645/72 & 102 &  10,320& 2,580& 1,440
		\\ \hline		
	\end{tabular}}
\caption{Detailed statistics of datasets.}

\label{tab:dataset}
\end{table}

\subsubsection{Metrics.}
For GZSL, the harmonic mean $H=(2MCA_u\times MCA_s)/(MCA_u+MCA_s)$ is widely used to evaluate the model performance, where $MCA_s$ and $MCA_u$ are the Mean Class Top-1 Accuracy for seen and unseen domains, respectively.

\subsection{Ablation Study}
In this part, we analyze each component of DCEN on CUB and aPY datasets for fast validation.
The baseline model is denoted as Basic-ZSL, which only learns $f(\cdot)$ via $\mathcal{L}_{ZSL}$ in Eq.~\eqref{eq:L_zsl} and is test via Eq.~\eqref{eq:infer}.

\begin{figure}[t]
	\centering
	\includegraphics[width=1\columnwidth]{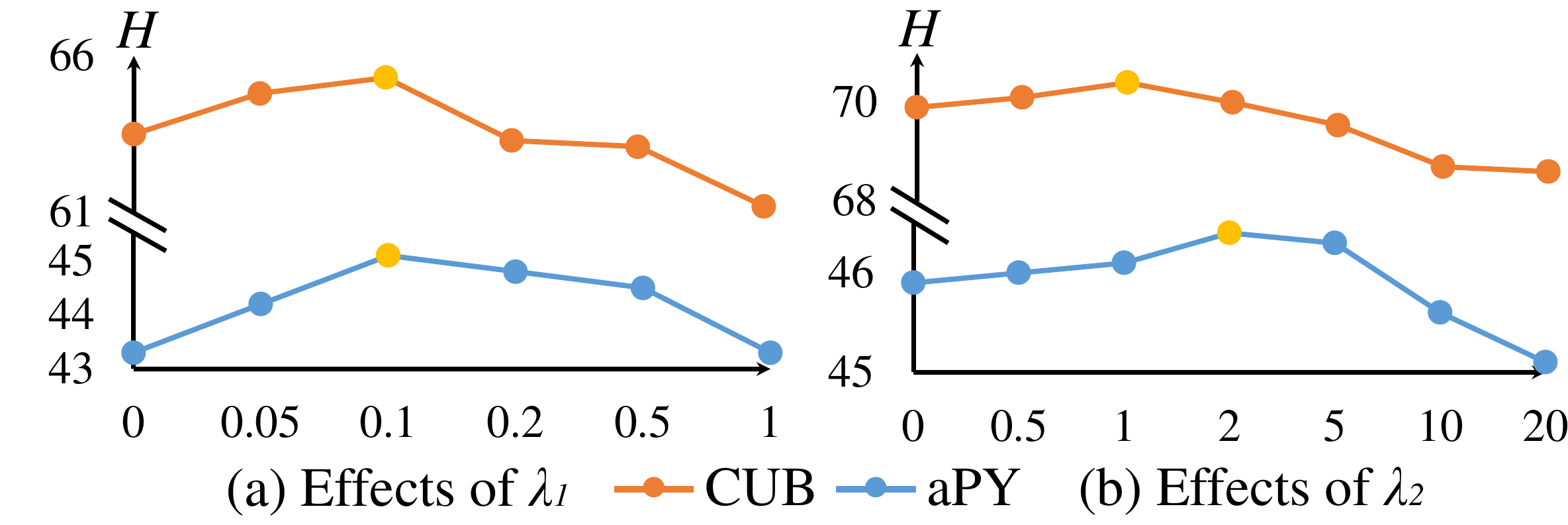} 
	
	\caption{Effects of different $\lambda_1$ and $\lambda_2$ for $\mathcal{L}_{id}$ and $\mathcal{L}_{sp}$, respectively. }
		\vspace{-0.5cm} 
	\label{fig:lambda}
\end{figure}

\subsubsection{Effects of instance discrimination.} The task-independent knowledge from instance discrimination is a major contribution of this paper, which is learned by Visual Contrastive Module (VCM).
Thus, we first analyze how the task-independent knowledge boosts the transferability of semantic-aligned representation in GZSL by varying $\lambda_1$ to influence the visual contrastive loss $\mathcal{L}_{id}$.
The results are given in Fig.~\ref{fig:lambda} (a).
In this experiment, we add VCM to Basic-ZSL with varying $\lambda_1$, thus $\lambda_1=0$ denotes no task-independent knowledge introduced. 
We can see that when increasing $\lambda_1$ from $0$ to $0.1$ for CUB and aPY, $\mathcal{L}_{id}$ begins introducing instance discriminative knowledge, and $H$ gradually increases. 
The best recognition performance is obtained at around $\lambda_1=0.1$ for both CUB and aPY.
This demonstrates that the task-independent knowledge brought by instance discrimination $\mathcal{L}_{id}$ enables DCEN to produce more transferable representations, thereby recognizing unseen domain images more accurately. 
When $\lambda_1$ increases from $0.1$ to $1.0$, the performance $H$ drops. 
Specifically, for CUB, the result with $\lambda_1=1$ is even worse than that with $\lambda_1=0$.
This reveals that, when $\lambda_1$ is large, the instance discrimination knowledge $\mathcal{L}_{id}$, which targets pushing all representations far away, becomes too sensitive to intra-class variance of different images and hard to preserve the semantic relationship, thereby being unable to cluster images correctly.
Thus, a trade-off between task-independent knowledge and task-specific knowledge is critical to make for strong transferable representations in GZSL.
We find $\lambda_1=0.1$ is suitable for most cases, which is used for the following experiments.

\begin{table}
	\centering	
	\linespread{0.87}
	\huge
	\resizebox{1\columnwidth}{!}{ 
	\begin{tabular}{l|c|c|c|c|c|c|c|c}
		\hline
		Aug.&Crop & Flip & Gray & CJ &Blur & Rotation & Swap   & \multirow{2}{*}{H} \\ \cline{1-8}
		prob.&1.0 & 0.5 & 0.2 & 0.8 & 0.5 & 0.5 & 0.2   &   \\ 
		\hline
		& &  &  &  &  &  &  & 65.3   \\ \cline{2-9}
		\multirow{16}{*}{\rotatebox{0}{$\mathcal{L}_{id}$}}&\cmark &  &  &  &  &  &  & 65.8$\uparrow$  \\ \cline{2-9}
		&\cmark & \cmark &  &  &  &  &    &66.0$\uparrow$  \\ \cline{2-9}
		&\cmark & \cmark & \cmark &  &  &  &    & 65.5$\downarrow$  \\ \cline{2-9}
		&\cmark & \cmark &  & \cmark(v1) &  &  &    & 63.8$\downarrow$ \\ \cline{2-9}
		&\cmark & \cmark &  & \cmark(v2) &  &  &    & 65.3$\downarrow$ \\ \cline{2-9}
		&\cmark & \cmark &  & \cmark(v3) &  &  &    & 65.2$\downarrow$ \\ \cline{2-9}
		&\cmark & \cmark &  &  & \cmark &  &    & 67.1$\uparrow$  \\ \cline{2-9}
		&\cmark & \cmark &  &  &  & \cmark(90) &    & 64.4$\downarrow$  \\ \cline{2-9}
		&\cmark & \cmark &  &  &  & \cmark(60) &    & 64.7$\downarrow$  \\ \cline{2-9}
		&\cmark & \cmark &  &  &  & \cmark(30) &    & 66.5$\uparrow$  \\ \cline{2-9}
		&\cmark & \cmark &  &  &  &  &  \cmark(7)  & 64.2$\downarrow$ \\ \cline{2-9}
		&\cmark & \cmark &  &  &  &  &  \cmark(5)  & 65.8$\downarrow$  \\ \cline{2-9}
		&\cmark & \cmark &  &  &  &  &  \cmark(3)  & 66.7$\uparrow$  \\ \cline{2-9}
		&\cmark & \cmark &  &  & \cmark & \cmark(30) &   &67.9$\uparrow$ \\ \cline{2-9}
		&\cmark & \cmark &  &  & \cmark & \cmark(30) & \cmark(3)   & \textbf{68.5}$\uparrow$  \\ \hline

	\end{tabular}}
\caption{Evaluating different visual augmentations on CUB by successively adding operations. When a certain operation brings positive effects, it is retained, otherwise, it is removed. The operation probability is experimentally determined. For v1, v2, and v3 color jittering, the parameters are respectively (0.4, 0.4, 0.4, 0.4), (0.4, 0.4, 0.4, 0.1), and (0.8, 0.8, 0.8, 0.2) for brightness, contrast, saturation, and hue. For rotation and swap, the parameters in parentheses indicate rotation angle and jigsaw number, respectively.}
\label{tab:visual augmentation}
\vspace{-0.5cm} 
\end{table}

\subsubsection{Effects of different invariance concepts.} In the task-independent knowledge brought by instance discrimination, the low-level invariance concepts, induced by visual augmentations, are important.
In this part, we conduct extensive experiments to explore useful invariance concepts in GZSL by combining different augmentations as $\varphi_{v}(\cdot)$ in VCM on CUB.
The experiments are carried out based on Basic-ZSL with $\lambda_1=0.1$.
The results are listed in Table \ref{tab:visual augmentation}, where we explore seven widely-used low-level augmentations, which contain random cropping, horizontal flipping, grayscale, color jittering, Gaussian blurring, rotation, and swapping.
Specifically, the swapping augmentation \cite{chen2019destruction} regards an image as a jigsaw and randomly shuffles the patches, which requires the representation to be invariant to patch-swapped images.
From the results, adding cropping, horizontal flipping, Gaussian blurring, rotation, and swap augmentations into $\varphi_{v}(\cdot)$ can benefit the performance, while adding grayscale and color jittering harms the results.
This is because that the grayscale and color augmentations will change the color space severely, which may prevent the model from inferring correct color attributes,~\emph{e.g.,} wing color, to recognize unseen categories.
Among the useful augmentations, Gaussian blurring brings the largest improvement because it helps the model neglect high-frequent noises.
For random rotation and swapping, too large angles and jigsaw numbers bring negative effects, which may destroy the image contents severely.
Finally, we combine all positive low-level augmentations in Table~\ref{tab:visual augmentation} as $\varphi_{v}(\cdot)$, which obtains an impressive improvement of $3.2\%$ over the baseline. The useful low-level invariance concepts explored in this experiment may provide an insight for the following GZSL studies.

 \begin{figure}[t]
	\centering
	\includegraphics[width=1\columnwidth]{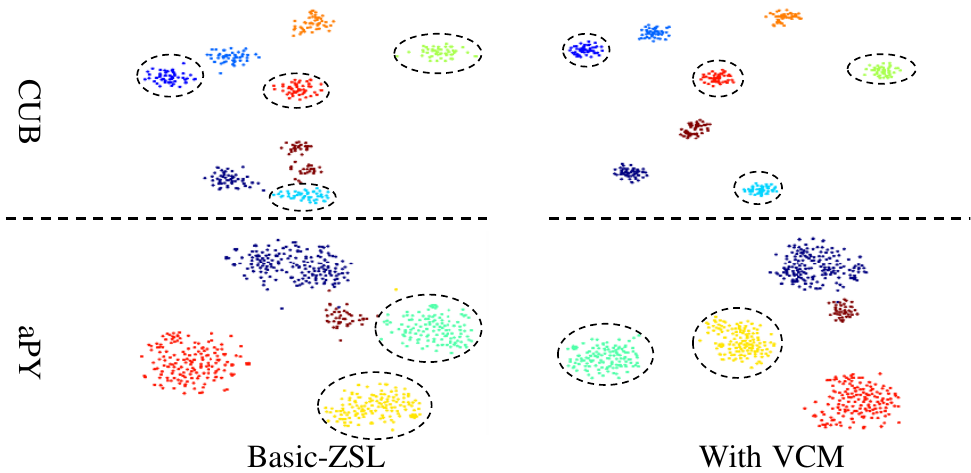} 
	
	\caption{Visualization of representation on CUB and aPY. Dotted circles denote unseen domain representations. }
	\vspace{-0.5cm} 
	\label{fig:vis}
\end{figure}

\subsubsection{Visualization of representation.}
To intuitively analyze the effect of task-independent knowledge from instance discrimination, we visualize the image representations before and after adding VCM using t-SNE \cite{van2008visualizing} on randomly selected categories.
The results are given in Fig.~\ref{fig:vis}.
After introducing task-independent knowledge into image representations, the category margin is obviously enlarged between seen and unseen categories, and the distribution of each individual cluster becomes more compact.
This means that image representations are more distinguishable between categories of the two domains, thus the unseen image representation is less biased towards the seen categories.
In other words, the visual representation can be better transferred to represent unseen image characteristics.
Besides, Table~\ref{tab:scm} gives quantitative results after adding VCM.
It shows that VCM can bring an impressive gain of $5\%$ on $H$ and $6.1\%$ on $MCA_u$, which demonstrates that VCM helps the model better transfer to the unseen domain.
In summary, both quantitative and qualitative results prove that the task-independent knowledge from instance discrimination can improve the representation transferability in GZSL.

\begin{table}
	\centering
	
	\begin{tabular}{c|c|c|c}
		\hline
		Methods&\textit{MCA}$_u$&\textit{MCA}$_s$&$H$
		\\ \hline
		Basic-ZSL & 56.3  & 72.8 & 63.5\\
		Basic-ZSL with VCM & 62.4  & 75.9 & 68.5\\
		+attribute masking & 62.5 & 78.3  & 69.5 \\
		+cross-modal triplet & 63.5 & 77.7 & 69.9 \\
		+$\mathcal{L}_{sp}$ & 63.8 & 78.4 & \textbf{70.4} \\
		\hline		
	\end{tabular}
\caption{Effects of each component of DCEN on CUB.}
\label{tab:scm}
\end{table}

 \begin{figure}[t]
 	\centering
 	\includegraphics[width=1\columnwidth]{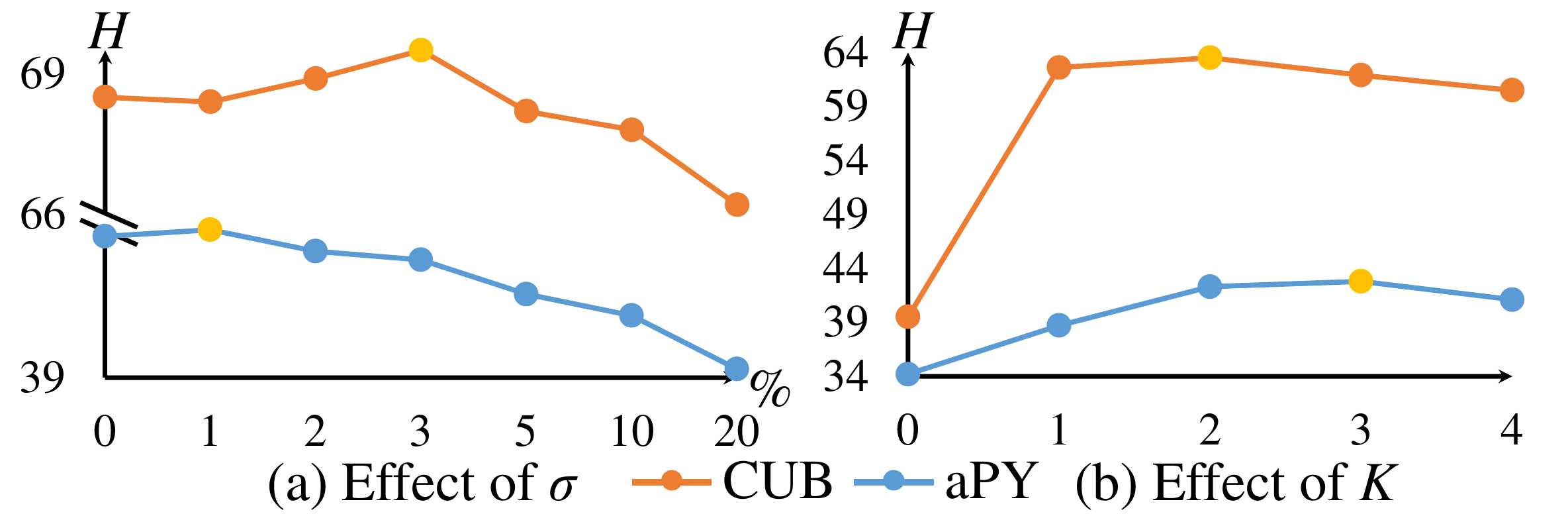} 
 	
 	\caption{Effects of different $\sigma$ and $K$. }
 	
 	\label{fig:K}
 	\vspace{-0.5cm} 
 \end{figure}

\begin{table*}
	\begin{center}
		 \linespread{0.9}
		\Huge
		\resizebox{2\columnwidth}{!}{ 
			\begin{tabular}{l|l|c|c|c|c|c|c|c|c|c|c|c|c}
				\hline
				&\multirow{2}{*}{Methods}&\multicolumn{3}{c|}{CUB~}&\multicolumn{3}{c|}{AWA2}&\multicolumn{3}{c|}{aPY}&\multicolumn{3}{c}{SUN}\\
				\cline{3-14}
				&&\textit{MCA}$_u$&\textit{MCA}$_s$&$H$&\textit{MCA}$_u$&\textit{MCA}$_s$&$H$&\textit{MCA}$_u$&\textit{MCA}$_s$&$H$&\textit{MCA}$_u$&\textit{MCA}$_s$&$H$\\
				\hline
				\hline
				\multirow{7}{*}{\rotatebox{270}{GEN}}&FGN\cite{xian2018feature}&43.7&57.7&49.7&-&-&-&-&-&-&42.6&36.6&39.4\\
				
				&SABR-I\cite{paul2019semantically}&55.0&58.7&56.8&30.3&93.9&46.9&-&-&-&50.7&35.1&41.5\\
				&f-VAEGAN-D2\cite{Xian_2019_CVPR}&63.2&75.6&68.9&-&-&-&-&-&-&50.1&37.8&43.1\\
				
				& RFF-GZSL\cite{Han_2020_CVPR} & 59.8 & 79.9 &  68.4 &-&-&-&-&-&-& 58.8 &  45.3 &  51.2\\
				
				& OCD-CVAE\cite{Keshari_2020_CVPR}&44.8 & 59.9 &  51.3 & 59.5 & 73.4 &  65.7&-&-&- & 44.8 & 42.9&  43.8\\
				& IZF-Softmax\cite{shen2020invertible} & 52.7 &68.0 &  59.4 & 60.6 & 77.5&  68.0 & 42.3 &60.5&  49.8& 52.7 & 57.0 & 54.8\\
				& TF-VAEGAN\cite{narayan2020latent} & 63.8 & 79.3 & 70.7 &-&-&-&-&-&-&  62.4 &47.1& 53.7\\
				
				\hline
				\hline
				\multirow{13}{*}{\rotatebox{270}{NON-GEN}}&SYNC\cite{Changpinyo2016}&11.5&70.9&19.8&10.0&90.5&18.0&7.4&66.3&13.3&7.9&43.3&13.4\\
				&CDL\cite{Jiang_2018_ECCV}&23.5&55.2&32.9&-&-&-&19.8&48.6&28.1&21.5&34.7&26.5\\
				&PSR-ZSL\cite{Annadani2018}&24.6&54.3&33.9&20.7&73.8&32.2&13.5&51.4&21.4&20.8&37.2&26.7\\
				&SP-AEN\cite{Chen2018}&34.7&70.6&46.6&23.3&90.9&37.1&13.7&63.4&22.6&24.9&38.6&30.3\\
				&DLFZRL\cite{tong2019hierarchical}&-&-&37.1&-&-&45.1&-&-&31.0&-&-&24.6\\
				&MLSE\cite{ding2019marginalized}&22.3&71.6&34.0&23.8&83.2&37.0&12.7&74.3&21.7&20.7&36.4&26.4\\
				&TripletLoss\cite{Cacheux_2019_ICCV}&55.8&52.3&53.0&48.5&83.2&61.3&-&-&-&47.9&30.4&36.8\\
				&COSMO\cite{atzmon2019adaptive}&44.4&57.8&50.2&-&-&-&-&-&-&44.9&37.7&\underline{41.0}\\
				&PREN\cite{ye2019progressive}&32.5&55.8&43.1&32.4&88.6&47.4&-&-&-&35.4&27.2&30.8\\
				&VSE-S\cite{zhu2019generalized} &33.4&87.5&48.4&41.6&91.3&57.2&24.5&72.0&36.6&-&-&-\\
				&AREN\cite{xie2019attentive}&63.2&69.0&\underline{66.0}&54.7&79.1&64.7&30.0&47.9&36.9&40.3&32.3&35.9\\
				&CosineSoftmax\cite{kampffmeyer2019rethinking}&47.4&47.6&47.5&56.4&81.4&66.7&26.5&74.0&\underline{39.0}&36.3&42.8&39.3\\
				
				& DAZLE\cite{Huynh_2020_CVPR} & 56.7&59.6 & 58.1 & 60.3&75.7 &\underline{67.1} & -& -& -&  52.3 & 24.3& 33.2\\
				\cline{2-14}
				
				&\textbf{DCEN} & 63.8 &78.4&\textbf{70.4}& 62.4 & 81.7 &  \textbf{70.8}  & 37.5 & 61.6 & \textbf{46.7}  & 43.7 & 39.8 &  \textbf{41.7 }\\
				\hline
		\end{tabular}}
	\caption{Results of GZSL on four classification benchmarks. Generative methods (GEN) utilize extra synthetic unseen domain data for training. The best result is bolded, and the second best is underlined. }
	\label{tab:gzsl}
	\end{center}
	\vspace{-0.5cm} 
\end{table*}
 
\subsubsection{Analysis of semantic contrastive module.} 
In this part, we analyze each component of SCM, and the results are given in Table~\ref{tab:scm}.
Based on Basic-ZSL with VCM, we add random attribute masking, cross-modal triplet loss, and masked attribute prediction sequentially.
First, we evaluate the effect of masking $\sigma\%$ attributes in SCM, which can improve model robustness to semantic noises. In training, attributes are randomly chosen with $p=0.5$, and $\sigma\%$ elements of these chosen attributes are masked. Random masking is not used in testing.  Fig.~\ref{fig:K} (a) shows the effect of varying $\sigma$ on CUB and aPY.
When $\sigma$ is too large which means many elements of attributes are missed, the semantic alignment becomes weak, thereby the performance drops.
Notably, the reason for small optimal $\sigma$ on aPY is that there are much fewer attributes in aPY than in CUB. Thus, suitable semantic masking is a simple way to improve semantic-visual robustness. 
Then, adding cross-modal triplet loss $\mathcal{L}_{sa}$ brings about $0.4\%$ improvement on CUB in Table~\ref{tab:scm}.
Specifically, $\mathcal{L}_{sa}$ contrasts cross-modal triplets to make visual representation not only close to corresponding semantic anchors but also far away from confusing negative categories, which improves representation discrimination.
Finally, we evaluate the effect of masked attribute prediction by varying $\lambda_2$ of $\mathcal{L}_{sp}$, and the results are given in Fig.~\ref{fig:lambda} (b).
We can see that setting appropriate weights of $\lambda_2$ for $\mathcal{L}_{sp}$ can boost the final results.
It proves that, by constraining the visual representation to predict the missed attributes, more semantic information from category attributes is explicitly encoded and preserved in visual representation.
In summary, our SCM can generate better semantic-aligned representation via cross-modal triplet loss and masked attribute prediction.

\subsubsection{Analysis for other hyper-parameters.} In DCEN, $K$, $\tau$, and $m$ are three minor hyper-parameters.
$K$ determines the architecture $h(\cdot)$ in Eq.~\eqref{eq:L_sa}.
Here, we evaluate the effects of different $K$ in Fig.~\ref{fig:K} (b).
It can be seen that $K=2$ is suitable for most cases.
$\tau$ and $m$ control the cosine similarity scaling in Eq.~\eqref{eq:L_id} and momentum updating of $g(\cdot)$ in Eq.~\eqref{eq:g_upate}.
In this paper, we set $\tau=0.07$ and $m=0.999$, which are commonly used in contrastive learning \cite{He_2020_CVPR}.

\subsection{Comparison with State-of-the-Art Methods}
Finally, we compare our DCEN with the state-of-the-art methods on four datasets,~\emph{i.e.,} CUB, AWA2, aPY, and SUN.
Results are given in Table~\ref{tab:gzsl}, which shows that DCEN surpasses existing methods on four datasets by a large margin.

First, we compare DCEN with related embedding-based methods, which use only seen domain data for training and no generative models.
DCEN surpasses the best related method by respectively $4.4\%$, $3.7\%$, $7.7\%$, $0.7\%$ on CUB, AWA2, aPY, and SUN datasets in the term of $H$, demonstrating its good generalizability to different image domains.
Notably, among non-generative methods, AREN \cite{xie2019attentive} obtains similar performance to DCEN, but it utilizes a model ensemble of two separate recognition branches.
Our improvements are impressive, because DCEN introduces extra task-independent knowledge into image representations via instance discrimination learning, and it is proved that appropriate task-independent knowledge can significantly improve representation transferability in GZSL.

Then, we compare DCEN with recent generative methods.
Notably, the generative methods utilize prior unseen domain semantics to synthesize extra unseen visual data for training,~\emph{e.g.,} powerful GANs, while DCEN only uses the seen domain data.
From Table~\ref{tab:gzsl}, DCEN outperforms most generative methods by a large margin, which is encouraging for embedding-based methods because no synthesized unseen domain data is used for training.
Compared to recent TF-VAEGAN~\cite{narayan2020latent}, DCEN obtains comparable results on CUB using only seen domain data.
This is due to introducing task-independent knowledge from instance discrimination, and it proves that the embedding-based methods have much potential.

Finally, we can conclude that: a) task-independent knowledge and task-specific semantic knowledge jointly make for strong transferable representations in GZSL, which enables DCEN to surpass all related works; and b) task-independent knowledge from instance discrimination is useful, and other task-independent knowledge remains to be explored.

\subsection{Discussion}
Compared to previous embedding-based methods, the improvement of DCEN on SUN is much less than the other three benchmarks.
The reason is that the number of images in each category of SUN is small,~\emph{e.g.,} averaged 16 images per category.
Thus, learning instance discrimination has a similar effect on image representation, compared to category discrimination.
This also reveals a limitation of DCEN,~\emph{i.e.,} the task-independent knowledge from instance discrimination may be not suitable for tail categories with inadequate samples.

\section{Conclusion}
In this paper, we propose a novel Dual-Contrastive Embedding Network (DCEN) that utilizes task-independent and task-specific knowledge to jointly make for transferable representation in GZSL. 
Specifically, a semantic contrastive module is developed to learn task-specific knowledge by performing cross-modal contrastive learning and exploring semantic-visual complementarity with category labels.
Besides, a visual contrastive module is designed to learn annotation-free task-independent knowledge via instance discrimination supervision, which gathers representations of the same image and pushes different image representations apart. 
Compared to seen category knowledge, the task-independent knowledge from instance discrimination is less biased, which can improve the representation transferability to unseen categories.
Extensive experiments show that our DCEN achieves superior performance on four GZSL benchmarks. 

In the future, the effects of more task-independent knowledge, such as rotation angle prediction and jigsaw order prediction, will be explored for GZSL. 

\section{Acknowledgements}
This work was supported by the National Key R\&D Program of China with grant No. 2020AAA0108602, National Natural Science Foundation of China (NSFC) under Grants 61632006 and 62076230, and Fundamental Research Funds for the Central Universities under Grants WK3490000003.

\section{Ethics Statement}

All datasets used in this paper are publicly available and have no relations with personal privacy information.

\bibliographystyle{aaai} 
\bibliography{aaai2}

\begin{thebibliography}{52}
\providecommand{\natexlab}[1]{#1}
\providecommand{\url}[1]{\texttt{#1}}
\providecommand{\urlprefix}{URL }
\expandafter\ifx\csname urlstyle\endcsname\relax
  \providecommand{\doi}[1]{doi:\discretionary{}{}{}#1}\else
  \providecommand{\doi}{doi:\discretionary{}{}{}\begingroup
  \urlstyle{rm}\Url}\fi

\bibitem[{Annadani and Biswas(2018)}]{Annadani2018}
Annadani, Y.; and Biswas, S. 2018.
\newblock Preserving Semantic Relations for Zero-Shot Learning.
\newblock In \emph{Proceedings of the IEEE Conference on Computer Vision and
  Pattern Recognition}, 7603--7612.

\bibitem[{Atzmon and Chechik(2019)}]{atzmon2019adaptive}
Atzmon, Y.; and Chechik, G. 2019.
\newblock Adaptive Confidence Smoothing for Generalized Zero-Shot Learning.
\newblock In \emph{Proceedings of the IEEE Conference on Computer Vision and
  Pattern Recognition}, 11671--11680.

\bibitem[{Ba et~al.(2015)Ba, Swersky, Fidler, and
  salakhutdinov}]{lei2015predicting}
Ba, J.~L.; Swersky, K.; Fidler, S.; and salakhutdinov, R. 2015.
\newblock Predicting Deep Zero-Shot Convolutional Neural Networks Using Textual
  Descriptions.
\newblock In \emph{Proceedings of the IEEE International Conference on Computer
  Vision}, 4247--4255.

\bibitem[{Badrinarayanan, Kendall, and
  Cipolla(2017)}]{badrinarayanan2017segnet}
Badrinarayanan, V.; Kendall, A.; and Cipolla, R. 2017.
\newblock Segnet: A Deep Convolutional Encoder-Decoder Architecture for Image
  Segmentation.
\newblock \emph{IEEE Transactions on Pattern Analysis and Machine Intelligence}
  39(12): 2481--2495.

\bibitem[{Cacheux, Borgne, and Crucianu(2019)}]{Cacheux_2019_ICCV}
Cacheux, Y.~L.; Borgne, H.~L.; and Crucianu, M. 2019.
\newblock Modeling Inter and Intra-Class Relations in the Triplet Loss for
  Zero-Shot Learning.
\newblock In \emph{Proceedings of the IEEE International Conference on Computer
  Vision}, 10333--10342.

\bibitem[{Caron et~al.(2018)Caron, Bojanowski, Joulin, and
  Douze}]{caron2018deep}
Caron, M.; Bojanowski, P.; Joulin, A.; and Douze, M. 2018.
\newblock Deep Clustering for Unsupervised Learning of Visual Features.
\newblock In \emph{Proceedings of the European Conference on Computer Vision},
  132--149.

\bibitem[{Changpinyo et~al.(2016)Changpinyo, Chao, Gong, and
  Sha}]{Changpinyo2016}
Changpinyo, S.; Chao, W.-L.; Gong, B.; and Sha, F. 2016.
\newblock Synthesized Classifiers for Zero-Shot Learning.
\newblock In \emph{Proceedings of the IEEE Conference on Computer Vision and
  Pattern Recognition}, 5327--5336.

\bibitem[{Chen et~al.(2018)Chen, Zhang, Xiao, Liu, and Chang}]{Chen2018}
Chen, L.; Zhang, H.; Xiao, J.; Liu, W.; and Chang, S.-F. 2018.
\newblock Zero-Shot Visual Recognition using Semantics-Preserving Adversarial
  Embedding Network.
\newblock In \emph{Proceedings of the IEEE Conference on Computer Vision and
  Pattern Recognition}, 1043--1052.

\bibitem[{Chen et~al.(2019)Chen, Bai, Zhang, and Mei}]{chen2019destruction}
Chen, Y.; Bai, Y.; Zhang, W.; and Mei, T. 2019.
\newblock Destruction and Construction Learning for Fine-Grained Image
  Recognition.
\newblock In \emph{Proceedings of the IEEE Conference on Computer Vision and
  Pattern Recognition}, 5157--5166.

\bibitem[{Cheng, Su, and Maji(2020)}]{cheng2020unsupervised}
Cheng, Z.; Su, J.-C.; and Maji, S. 2020.
\newblock Unsupervised Discovery of Object Landmarks via Contrastive Learning.
\newblock \emph{arXiv preprint arXiv:2006.14787} .

\bibitem[{Ding and Liu(2019)}]{ding2019marginalized}
Ding, Z.; and Liu, H. 2019.
\newblock Marginalized Latent Semantic Encoder for Zero-Shot Learning.
\newblock In \emph{Proceedings of the IEEE Conference on Computer Vision and
  Pattern Recognition}, 6191--6199.

\bibitem[{Farhadi et~al.(2009)Farhadi, Endres, Hoiem, and
  Forsyth}]{farhadi2009describing}
Farhadi, A.; Endres, I.; Hoiem, D.; and Forsyth, D. 2009.
\newblock Describing Objects by Their Attributes.
\newblock In \emph{Proceedings of the IEEE Conference on Computer Vision and
  Pattern Recognition}, 1778--1785.

\bibitem[{Felix et~al.(2018)Felix, Kumar, Reid, and Carneiro}]{felix2018multi}
Felix, R.; Kumar, V.~B.; Reid, I.; and Carneiro, G. 2018.
\newblock Multi-Modal Cycle-Consistent Generalized Zero-Shot Learning.
\newblock In \emph{Proceedings of the European Conference on Computer Vision},
  21--37.

\bibitem[{Gidaris et~al.(2019)Gidaris, Bursuc, Komodakis, P{\'e}rez, and
  Cord}]{gidaris2019boosting}
Gidaris, S.; Bursuc, A.; Komodakis, N.; P{\'e}rez, P.; and Cord, M. 2019.
\newblock Boosting Few-Shot Visual Learning with Self-Supervision.
\newblock In \emph{Proceedings of the IEEE International Conference on Computer
  Vision}, 8059--8068.

\bibitem[{Gidaris, Singh, and Komodakis(2018)}]{gidaris2018unsupervised}
Gidaris, S.; Singh, P.; and Komodakis, N. 2018.
\newblock Unsupervised Representation Learning by Predicting Image Rotations.
\newblock In \emph{International Conference on Learning Representations},
  1--16.

\bibitem[{Girshick et~al.(2014)Girshick, Donahue, Darrell, and
  Malik}]{girshick2014rich}
Girshick, R.; Donahue, J.; Darrell, T.; and Malik, J. 2014.
\newblock Rich Feature Hierarchies for Accurate Object Detection and Semantic
  Segmentation.
\newblock In \emph{Proceedings of the IEEE Conference on Computer Vision and
  Pattern Recognition}, 580--587.

\bibitem[{Han, Fu, and Yang(2020)}]{Han_2020_CVPR}
Han, Z.; Fu, Z.; and Yang, J. 2020.
\newblock Learning the Redundancy-Free Features for Generalized Zero-Shot
  Object Recognition.
\newblock In \emph{Proceedings of the IEEE Conference on Computer Vision and
  Pattern Recognition}, 12865--12874.

\bibitem[{He et~al.(2020)He, Fan, Wu, Xie, and Girshick}]{He_2020_CVPR}
He, K.; Fan, H.; Wu, Y.; Xie, S.; and Girshick, R. 2020.
\newblock Momentum Contrast for Unsupervised Visual Representation Learning.
\newblock In \emph{Proceedings of the IEEE Conference on Computer Vision and
  Pattern Recognition}, 9729--9738.

\bibitem[{He et~al.(2016)He, Zhang, Ren, and Sun}]{He_2016_CVPR}
He, K.; Zhang, X.; Ren, S.; and Sun, J. 2016.
\newblock Deep Residual Learning for Image Recognition.
\newblock In \emph{Proceedings of the IEEE Conference on Computer Vision and
  Pattern Recognition}, 770--778.

\bibitem[{Huynh and Elhamifar(2020)}]{Huynh_2020_CVPR}
Huynh, D.; and Elhamifar, E. 2020.
\newblock Fine-Grained Generalized Zero-Shot Learning via Dense Attribute-Based
  Attention.
\newblock In \emph{Proceedings of the IEEE Conference on Computer Vision and
  Pattern Recognition}, 4483--4493.

\bibitem[{Jiang et~al.(2018)Jiang, Wang, Shan, and Chen}]{Jiang_2018_ECCV}
Jiang, H.; Wang, R.; Shan, S.; and Chen, X. 2018.
\newblock Learning Class Prototypes via Structure Alignment for Zero-Shot
  Recognition.
\newblock In \emph{Proceedings of the European Conference on Computer Vision},
  118--134.

\bibitem[{Kampffmeyer et~al.(2019)Kampffmeyer, Chen, Liang, Wang, Zhang, and
  Xing}]{kampffmeyer2019rethinking}
Kampffmeyer, M.; Chen, Y.; Liang, X.; Wang, H.; Zhang, Y.; and Xing, E.~P.
  2019.
\newblock Rethinking Knowledge Graph Propagation for Zero-Shot Learning.
\newblock In \emph{Proceedings of the IEEE Conference on Computer Vision and
  Pattern Recognition}, 11487--11496.

\bibitem[{Keshari, Singh, and Vatsa(2020)}]{Keshari_2020_CVPR}
Keshari, R.; Singh, R.; and Vatsa, M. 2020.
\newblock Generalized Zero-Shot Learning via Over-Complete Distribution.
\newblock In \emph{Proceedings of the IEEE Conference on Computer Vision and
  Pattern Recognition}, 13300--13308.

\bibitem[{Kodirov, Xiang, and Gong(2017)}]{kodirov2017semantic}
Kodirov, E.; Xiang, T.; and Gong, S. 2017.
\newblock Semantic Autoencoder for Zero-Shot Learning.
\newblock In \emph{Proceedings of the IEEE Conference on Computer Vision and
  Pattern Recognition}, 3174--3183.

\bibitem[{Laine and Aila(2017)}]{laine2016temporal}
Laine, S.; and Aila, T. 2017.
\newblock Temporal Ensembling for Semi-supervised Learning.
\newblock In \emph{International Conference on Learning Representations},
  1--13.

\bibitem[{Liu et~al.(2019)Liu, Guo, Cai, and He}]{Liu_2019_ICCV}
Liu, Y.; Guo, J.; Cai, D.; and He, X. 2019.
\newblock Attribute Attention for Semantic Disambiguation in Zero-Shot
  Learning.
\newblock In \emph{Proceedings of the IEEE International Conference on Computer
  Vision}, 6698--6707.

\bibitem[{Min et~al.(2019)Min, Yao, Xie, Zha, and Zhang}]{min2019domain}
Min, S.; Yao, H.; Xie, H.; Zha, Z.-J.; and Zhang, Y. 2019.
\newblock Domain-Specific Embedding Network for Zero-Shot Recognition.
\newblock In \emph{Proceedings of ACM International Conference on Multimedia},
  2070--2078.

\bibitem[{Min et~al.(2020)Min, Yao, Xie, Zha, and Zhang}]{min2020domain}
Min, S.; Yao, H.; Xie, H.; Zha, Z.-J.; and Zhang, Y. 2020.
\newblock Domain-Oriented Semantic Embedding for Zero-Shot Learning.
\newblock \emph{IEEE Transactions on Multimedia} .

\bibitem[{Morgado and Vasconcelos(2017)}]{Morgado2017}
Morgado, P.; and Vasconcelos, N. 2017.
\newblock Semantically Consistent Regularization for Zero-Shot Recognition.
\newblock In \emph{Proceedings of the IEEE Conference on Computer Vision and
  Pattern Recognition}, 6060--6069.

\bibitem[{Morgado, Vasconcelos, and Misra(2020)}]{morgado2020audio}
Morgado, P.; Vasconcelos, N.; and Misra, I. 2020.
\newblock Audio-Visual Instance Discrimination with Cross-Modal Agreement.
\newblock \emph{arXiv preprint arXiv:2004.12943} .

\bibitem[{Narayan et~al.(2020)Narayan, Gupta, Khan, Snoek, and
  Shao}]{narayan2020latent}
Narayan, S.; Gupta, A.; Khan, F.~S.; Snoek, C.~G.; and Shao, L. 2020.
\newblock Latent Embedding Feedback and Discriminative Features for Zero-Shot
  Classification.
\newblock In \emph{Proceedings of the European Conference on Computer Vision},
  1--23.

\bibitem[{Noroozi and Favaro(2016)}]{noroozi2016unsupervised}
Noroozi, M.; and Favaro, P. 2016.
\newblock Unsupervised Learning of Visual Representations by Solving Jigsaw
  Puzzles.
\newblock In \emph{Proceedings of the European Conference on Computer Vision},
  69--84.

\bibitem[{Patterson and Hays(2012)}]{patterson2012sun}
Patterson, G.; and Hays, J. 2012.
\newblock Sun Attribute Database: Discovering, Annotating, and Recognizing
  Scene Attributes.
\newblock In \emph{Proceedings of the IEEE Conference on Computer Vision and
  Pattern Recognition}, 2751--2758.

\bibitem[{Paul, Krishnan, and Munjal(2019)}]{paul2019semantically}
Paul, A.; Krishnan, N.~C.; and Munjal, P. 2019.
\newblock Semantically Aligned Bias Reducing Zero Shot Learning.
\newblock In \emph{Proceedings of the IEEE Conference on Computer Vision and
  Pattern Recognition}, 7056--7065.

\bibitem[{Schroff, Kalenichenko, and Philbin(2015)}]{schroff2015facenet}
Schroff, F.; Kalenichenko, D.; and Philbin, J. 2015.
\newblock FaceNet: A Unified Embedding for Face Recognition and Clustering.
\newblock In \emph{Proceedings of the IEEE Conference on Computer Vision and
  Pattern Recognition}, 815--823.

\bibitem[{Shen, Qin, and Huang(2020)}]{shen2020invertible}
Shen, Y.; Qin, J.; and Huang, L. 2020.
\newblock Invertible Zero-Shot Recognition Flows.
\newblock In \emph{Proceedings of the European Conference on Computer Vision},
  614--631.

\bibitem[{Shigeto et~al.(2015)Shigeto, Suzuki, Hara, Shimbo, and
  Matsumoto}]{Shigeto2015}
Shigeto, Y.; Suzuki, I.; Hara, K.; Shimbo, M.; and Matsumoto, Y. 2015.
\newblock Ridge Regression, Hubness, and Zero-Shot Learning.
\newblock In \emph{Joint European Conference on Machine Learning and Knowledge
  Discovery in Databases}, 135--151.

\bibitem[{Tian, Krishnan, and Isola(2020)}]{tian2020contrastive}
Tian, Y.; Krishnan, D.; and Isola, P. 2020.
\newblock Contrastive Multiview Coding.
\newblock In \emph{Proceedings of the European Conference on Computer Vision},
  776--794.

\bibitem[{Tong et~al.(2019)Tong, Wang, Klinkigt, Kobayashi, and
  Nonaka}]{tong2019hierarchical}
Tong, B.; Wang, C.; Klinkigt, M.; Kobayashi, Y.; and Nonaka, Y. 2019.
\newblock Hierarchical Disentanglement of Discriminative Latent Features for
  Zero-Shot Learning.
\newblock In \emph{Proceedings of the IEEE Conference on Computer Vision and
  Pattern Recognition}, 11467--11476.

\bibitem[{Van~der Maaten and Hinton(2008)}]{van2008visualizing}
Van~der Maaten, L.; and Hinton, G. 2008.
\newblock Visualizing data using t-SNE.
\newblock \emph{Journal of Machine Learning Research} 9(11): 2579–2605.

\bibitem[{Wah et~al.(2011)Wah, Branson, Welinder, Perona, and
  Belongie}]{Welinder2010}
Wah, C.; Branson, S.; Welinder, P.; Perona, P.; and Belongie, S. 2011.
\newblock The Caltech-UCSD Birds-200-2011 Dataset.
\newblock In \emph{California Institute of Technology}, 1--8.

\bibitem[{Wu et~al.(2020)Wu, Zhang, Zha, Luo, Zhang, and Wu}]{wu2020self}
Wu, J.; Zhang, T.; Zha, Z.-J.; Luo, J.; Zhang, Y.; and Wu, F. 2020.
\newblock Self-Supervised Domain-Aware Generative Network for Generalized
  Zero-Shot Learning.
\newblock In \emph{Proceedings of the IEEE Conference on Computer Vision and
  Pattern Recognition}, 12767--12776.

\bibitem[{Xian et~al.(2016)Xian, Akata, Sharma, Nguyen, Hein, and
  Schiele}]{Xian2016}
Xian, Y.; Akata, Z.; Sharma, G.; Nguyen, Q.; Hein, M.; and Schiele, B. 2016.
\newblock Latent Embeddings for Zero-Shot Classification.
\newblock In \emph{Proceedings of the IEEE Conference on Computer Vision and
  Pattern Recognition}, 69--77.

\bibitem[{Xian et~al.(2018{\natexlab{a}})Xian, Lampert, Schiele, and
  Akata}]{xian2018zero}
Xian, Y.; Lampert, C.~H.; Schiele, B.; and Akata, Z. 2018{\natexlab{a}}.
\newblock Zero-shot learning—A Comprehensive Evaluation of the Good, the Bad
  and the Ugly.
\newblock \emph{IEEE Transactions on Pattern Analysis and Machine Intelligence}
  41(9): 2251--2265.

\bibitem[{Xian et~al.(2018{\natexlab{b}})Xian, Lorenz, Schiele, and
  Akata}]{xian2018feature}
Xian, Y.; Lorenz, T.; Schiele, B.; and Akata, Z. 2018{\natexlab{b}}.
\newblock Feature Generating Networks for Zero-Shot Learning.
\newblock In \emph{Proceedings of the IEEE Conference on Computer Vision and
  Pattern Recognition}, 5542--5551.

\bibitem[{Xian et~al.(2019)Xian, Sharma, Schiele, and Akata}]{Xian_2019_CVPR}
Xian, Y.; Sharma, S.; Schiele, B.; and Akata, Z. 2019.
\newblock F-VAEGAN-D2: A Feature Generating Framework for Any-Shot Learning.
\newblock In \emph{Proceedings of the IEEE Conference on Computer Vision and
  Pattern Recognition}, 10275--10284.

\bibitem[{Xie et~al.(2019)Xie, Liu, Jin, Zhu, Zhang, Qin, Yao, and
  Shao}]{xie2019attentive}
Xie, G.-S.; Liu, L.; Jin, X.; Zhu, F.; Zhang, Z.; Qin, J.; Yao, Y.; and Shao,
  L. 2019.
\newblock Attentive Region Embedding Network for Zero-shot Learning.
\newblock In \emph{Proceedings of the IEEE Conference on Computer Vision and
  Pattern Recognition}, 9384--9393.

\bibitem[{Yang et~al.(2018)Yang, Deng, Liu, and Nie}]{yang2018new}
Yang, X.; Deng, C.; Liu, X.; and Nie, F. 2018.
\newblock New l2, 1-norm Relaxation of Multi-Way Graph Cut for Clustering.
\newblock In \emph{AAAI Conference on Artificial Intelligence}, 1--8.

\bibitem[{Yang et~al.(2020)Yang, Deng, Wei, Yan, and Liu}]{yang2020adversarial}
Yang, X.; Deng, C.; Wei, K.; Yan, J.; and Liu, W. 2020.
\newblock Adversarial Learning for Robust Deep Clustering.
\newblock In \emph{Advances in Neural Information Processing Systems},
  9098--9108.

\bibitem[{Ye and Guo(2019)}]{ye2019progressive}
Ye, M.; and Guo, Y. 2019.
\newblock Progressive Ensemble Networks for Zero-Shot Recognition.
\newblock In \emph{Proceedings of the IEEE Conference on Computer Vision and
  Pattern Recognition}, 11728--11736.

\bibitem[{Zhu, Wang, and Saligrama(2019)}]{zhu2019generalized}
Zhu, P.; Wang, H.; and Saligrama, V. 2019.
\newblock Generalized Zero-Shot Recognition based on Visually Semantic
  Embedding.
\newblock In \emph{Proceedings of the IEEE Conference on Computer Vision and
  Pattern Recognition}, 2995--3003.

\bibitem[{Zhu et~al.(2019)Zhu, Xie, Tang, Peng, and Elgammal}]{zhu2019semantic}
Zhu, Y.; Xie, J.; Tang, Z.; Peng, X.; and Elgammal, A. 2019.
\newblock Semantic-Guided Multi-Attention Localization for Zero-Shot Learning.
\newblock In \emph{Advances in Neural Information Processing Systems}, 1--11.

\end{thebibliography}

\end{document}